\documentclass[preprint,12pt,authoryear]{elsarticle}

\usepackage{amssymb}
\usepackage{amsmath}

\usepackage{graphicx}%
\usepackage{multirow}%
\usepackage{amsmath,amssymb,amsfonts}%
\usepackage{amsthm}%
\usepackage{mathrsfs}%
\usepackage[title]{appendix}%
\usepackage{xcolor}%
\usepackage{textcomp}%
\usepackage{manyfoot}%
\usepackage{booktabs}%
\usepackage{algorithm}%
\usepackage{algorithmicx}%
\usepackage{algpseudocode}%
\usepackage{listings}%
\usepackage{paralist}%
\usepackage{adjustbox}%
\usepackage{hyperref}

\hypersetup{
    colorlinks=true,    
    linkcolor=blue,   
    citecolor=blue,    
    filecolor=blue,     
    urlcolor=blue   
}

\journal{Artificial Intelligence Journal}

\begin{document}

\begin{frontmatter}


\title{Can LLM-Augmented autonomous agents cooperate?, An evaluation of their cooperative capabilities through Melting Pot}

\cortext[label_contributing]{Contributing author}
\author[DISC]{Manuel Mosquera}
\ead{ma.mosquerao@uniandes.edu.co}
\author[DISC]{Juan Sebastian Pinzón}
\ead{js.pinzonr@uniandes.edu.co}
\author[IELE]{Yesid Fonseca}
\ead{y.fonseca@uniandes.edu.co}
\author[BANCOL]{Manuel Ríos}
\ead{manrios@bancolombia.com.co}
\author[IELE]{Nicanor Quijano}
\ead{nquijano@uniandes.edu.co}
\author[IBIO]{Luis Felipe Giraldo}
\ead{lf.giraldo404@uniandes.edu.co}
\author[DISC]{Rubén Manrique\corref{label_contributing}}
\ead{rf.manrique@uniandes.edu.co}
\affiliation[DISC]{organization={Department of Systems and Computing Engineering, Los Andes University},
           addressline={address}, 
           city={Bogotá},
           postcode={111711}, 
           country={Colombia}}
\affiliation[IBIO]{organization={Department of Biomedical Engineering, Los Andes University},
           addressline={address}, 
           city={Bogotá},
           postcode={111711}, 
           country={Colombia}}
\affiliation[IELE]{organization={Department of Electrical and Electronic Engineering, Los Andes University},
           addressline={address}, 
           city={Bogotá},
           postcode={111711}, 
           country={Colombia}}
\affiliation[BANCOL]{organization={Center of Excellence in Analytics and Artificial Intelligence, Bancolombia},
           city={Bogotá},
           postcode={111711}, 
           country={Colombia}}





\begin{abstract}
As the field of AI continues to evolve, a significant dimension of this progression is the development of Large Language Models (LLMs) and their potential to enhance multi-agent artificial intelligence systems. This paper explores the cooperative capabilities of Large Language Model-augmented Autonomous Agents (LAAs) using the well-known Melting Pot environments along with reference models such as GPT-4 and GPT-3.5. Preliminary results suggest that while these agents demonstrate a propensity for cooperation, they still struggle with effective collaboration in given environments, emphasizing the need for more robust architectures. The study's contributions include an abstraction layer to adapt Melting Pot game scenarios for LLMs, the implementation of a reusable architecture for LLM-mediated agent development (which includes short and long-term memories and different cognitive modules), and the evaluation of cooperation capabilities using a set of metrics tied to the Melting Pot's ``Commons Harvest'' game. The paper closes by discussing the limitations of the current architectural framework and the potential of a new set of modules that fosters better cooperation among LAAs. 
\end{abstract}


\end{frontmatter}


\section{Introduction}\label{sec1}

The increased presence and relevance of AI agents within everyday spheres such as self-driving vehicles and customer service necessitates, these entities being equipped with the appropriate capabilities to facilitate cooperation with humans and its AI counterparts. While noteworthy strides have been made in advancing individual intelligence components within AI agents, expanding the research focus to enhance their social intelligence ---the ability to effectively collaborate within group settings to solve prevalent problems–- is now timely. This pivot aligns with the rapid progression of AI research presenting fresh prospects for fostering cooperation, drawing on insights from social choice theory and the development of social systems \citep{openproblems}.

Moreover, using AI to manage open innovation processes provides a framework for improving AI's social intelligence. By integrating AI to perform key functions such as mapping the innovation landscape, coordinating diverse knowledge inputs, and ensuring collaboration aligns with collective objectives, we can better understand and leverage AI's capabilities in facilitating complex collaborative efforts within and across communities \citep{openinnovation}.

Cooperation requires the presence of two or more agents who, based on mutual understanding, engage in collaborative actions. From the perspective of evolution, cooperation has played a crucial role in the survival of species \citep{Pennisi2009,Dale2020}. It has facilitated the development of social structures that influence our surroundings and has provided solutions to complex issues, such as social dilemmas \citep{Gross2023}. Unveiling the processes that enable the evolution of cooperative behavior in communities is regarded as an important challenge that scientists should tackle \citep{Pennisi2009}.

Mathematical and computational models of cooperation have been proposed to study the mechanisms that facilitate the emergence of cooperation among artificial agents engaged in collaborative actions with the aim of improving their joint welfare \citep{openproblems}. For example, matrix games have provided a useful tool to conduct research on social dilemmas for years \citep{Axelrod1981}, in which the decision to either cooperate or defect is restricted to an atomic action. Additionally, multi-agent reinforcement learning has been used to study the process of learning cooperative policies in complex scenarios where temporally extended social dilemmas arise \citep{Leibo2017, leibo2021scalable, McKee2023, rios2023understanding}.

Research into AI agents presents an avenue for generating intelligent technologies that embody more human-like features and are compatible with humans, a far cry from solipsistic approaches that overlook agent interactions. A promising illustration of this approach is the Melting Pot, an AI research tool designed to foster collaborative efforts within multi-agent artificial intelligence via canonical test scenarios. These environments emphasize non-trivial, learnable, and measurable cooperation by pairing a physical environment (a ``substrate'') with a reference set of co-players (a ``background population'') \citep{meltingpot}. The environments fostered interdependence between the individuals involved.

Research on AI agents has also recently been permeated by the leaps and bounds of Large Language Models (LLMs). The increasing success of LLMs encourages further exploration into LLM-augmented Autonomous Agents (LAAs).  LAAs represent an avenue of research that is still emerging, with limited explorations currently available \citep{MultiAgentDebate, bolaa, collaborationpsychology, reflexion, toolformer, react}. Something common in these works is that a clear need is established. To achieve success, LAAs have to rely on an architecture that can recall relevant events, reflect on such memories to generalize and draw a higher level of inferences, and utilize those reasonings to develop timely and long-term plans \citep{generativeagents}. 

These architectures offer specialized modules for specific tasks, utilizing meticulously crafted prompts and flows to perform complicated tasks and navigate intricate environments. Human behavior replication has been observed in some of these architectures, notably by  \citet{generativeagents}, who managed to create convincingly realistic human behavior in simulated environments. Similar frameworks utilized by Voyager \citep{voyager} enabled an agent to navigate the Minecraft world and independently develop tools and skills. MetaGPT \citep{metagpt} introduced a framework allowing for the generation of fully functional programs, simulating a business-like environment with distinct roles and predefined agent interactions.

Despite significant advancements in the field, the potential for cooperative abilities in Large Language Models augmented Autonomous Agents (LAAs) has been somewhat neglected in current research. These capabilities could, however, be paramount in empowering these agents to perform innovative tasks and succeed in complex environments. This study represents an initial exploration into the inherent cooperative capabilities of LAAs. We employ an evaluation framework that includes a communication interface of scenarios from the Melting Pot project \citep{meltingpot} (in which artificial agents co-exist in environments where social dilemmas can arise), the recent architecture proposed by  \citet{generativeagents}, as well as reference Large Language Models (LLMs) such as GPT-4 and GPT-3.5. Our results hint towards the capability for cooperative behavior, based on simple natural language definitions and cooperation metrics tailored to the chosen Melting Pot scenario. While the agents showed a propensity to cooperate, their actions did not demonstrate a clear understanding of effective collaboration within the given environment. Consequently, our analysis underscores the necessity for more robust architectures that can foster better collaboration in LAAs.

In summary, our contributions are as follows:

\begin{itemize}
    \item Adapting the Melting Pot scenarios to textual representations that can be easily operationalized by LLMs.
    \item Implementing a reusable architecture for the development of LAAs employing the modules proposed in Generative Agents \citep{generativeagents}. This architecture includes short- and long-term memories and cognitive modules of perception, planning, reflection, and action. Our project can be found at \href{https://github.com/Cooperative-IA/CooperativeGPT}{https://github.com/Cooperative-IA/CooperativeGPT}
    \item Implementing ``personalities'' specified in natural language, making it clear to the agents whether they should be cooperative or not. These descriptions are intended to discern, based on their pre-training knowledge, what they perceive as cooperation in an unfamiliar context.
    \item Evaluating LLM-mediated agents in the ``Commons Harvest'' game of Melting Pot using our architecture in different scenarios where we specify or not, through natural language, the personality of the agents.
    \item Discussing the results in terms of cooperativity metrics associated with the ``Commons Harvest'' game, the limitations of the used architecture, and the proposal of an improved architecture that fosters better cooperation among LAAs.
\end{itemize}

\section{Related Work}\label{sec2}

Agent architectures have evolved to address the limitations of traditional LLMs, equipping them with diverse tools for autonomous operation or minimal human oversight.

A notable challenge with LLMs is their susceptibility to hallucinations and gaps in knowledge regarding recent events or specific subjects. Such constraints diminish their practicality, as they remain confined to the information acquired during training without the capability to assimilate new data. To address this issue,  \citet{toolformer} introduced an early solution named Toolformer. This model was trained to discern when and how to invoke APIs (tools) to enhance the LLM's performance across various tasks. The dataset was self-supervised, with API calls incorporated only when they positively impacted the model's performance. This methodology empowered the model to determine the relevance and optimal execution of API calls.

However, for executing more intricate tasks, merely invoking tools may be insufficient. Toolformer lacks the capability to reason about the rationale behind API calls and does not receive comprehensive environmental feedback to guide its subsequent actions toward achieving a goal. Recognizing this gap, the prompt-based paradigm ReAct, developed by  \citet{react}, integrates reasoning with action. By providing contextual prompt examples, ReAct guides the LLM on when to engage in reasoning and when to act, resulting in enhanced performance compared to approaches that employ reasoning or action in isolation. Furthermore, to enable the agent to learn from its errors,  \citet{reflexion} expanded the ReAct framework by incorporating a self-reflection module. This addition offers verbal feedback on past unsuccessful attempts, facilitating performance enhancement in subsequent trials.

Conversely, drawing inspiration from the emulation of authentic human behavior,  \citet{generativeagents} devised an intricate agent framework. This architecture boasts a cognitive sequence structured around modules primarily anchored by diverse prompts. Leveraging distinct prompts optimizes LLM performance, enabling specialized techniques for specific tasks. Notably, memory holds a pivotal position within this framework, preserving the agent's experiences and insights. The ability to retrieve these memories diversely amplifies their utility across multiple objectives. Moreover, \citet{voyager} developed the Voyager architecture, enabling autonomous gameplay in Minecraft. This advanced framework empowers the agent to autonomously curate a discovery agenda. Remarkably, the architecture can also create its own APIs, write corresponding code, verify API functionality, and store it in a vector database for future utilization.

More recently, efforts have been directed towards enhancing agent performance through multi-agent frameworks that utilize different instances of LLMs to independently perform roles or tasks. \citet{MultiAgentDebate} demonstrated this through a framework designed to engage different LLM instances in a debate, aiming to improve the factuality and accuracy of the responses. Subsequently, Hong et al.\ further capitalized on the potential of multiple agents. They allocated specific roles to each agent, accompanied by a sequence of predefined tasks with clear input and output expectations. These tasks establish a structured interaction pathway between agents, enabling them to achieve user-defined objectives. Demonstrating its efficacy in software-related tasks, this framework, inspired by the operational dynamics of conventional software companies, attained state-of-the-art performance in the HumanEval and MBPP benchmarks.

Similarly, \citet{bolaa} introduced the BOLAA framework. This system orchestrates multi-agent activity by defining specialized agents overseen by a central controller. The controller's role is pivotal: it selects the most suitable agent for a given task and facilitates communication with it. Additionally, \citet{collaborationpsychology} delved into multi-agent architectures, exploring the influence of social traits and collaborative strategies on different datasets. Likewise, \citet{mechagents} developed ``MechAgents,'' a system employing multiple dynamically interacting LLMs to autonomously solve mechanics tasks. This framework not only shows that self-correcting and mutually correcting code can work between AI agents, but it also makes it easier for the agents to combine physics-based modeling with domain-specific knowledge. This opens up a new way to automate and improve the way engineers solve problems.

\section{Methodology}\label{sec3}

\subsection{Experimental setup}\label{subsec31}

\subsubsection{Environment} This paper utilizes a scenario sourced from Melting Pot \citep{meltingpot}, a research tool developed by DeepMind for the purpose of experimentation and evaluation within the realm of multi-agent artificial intelligence. The scenarios within Melting Pot are specifically crafted to establish social situations in which the ability of the agents to solve conflict is challenged and characterized by significant interdependence among the involved agents.

\begin{figure}[h]%
\centering
\includegraphics[width=0.8\textwidth]{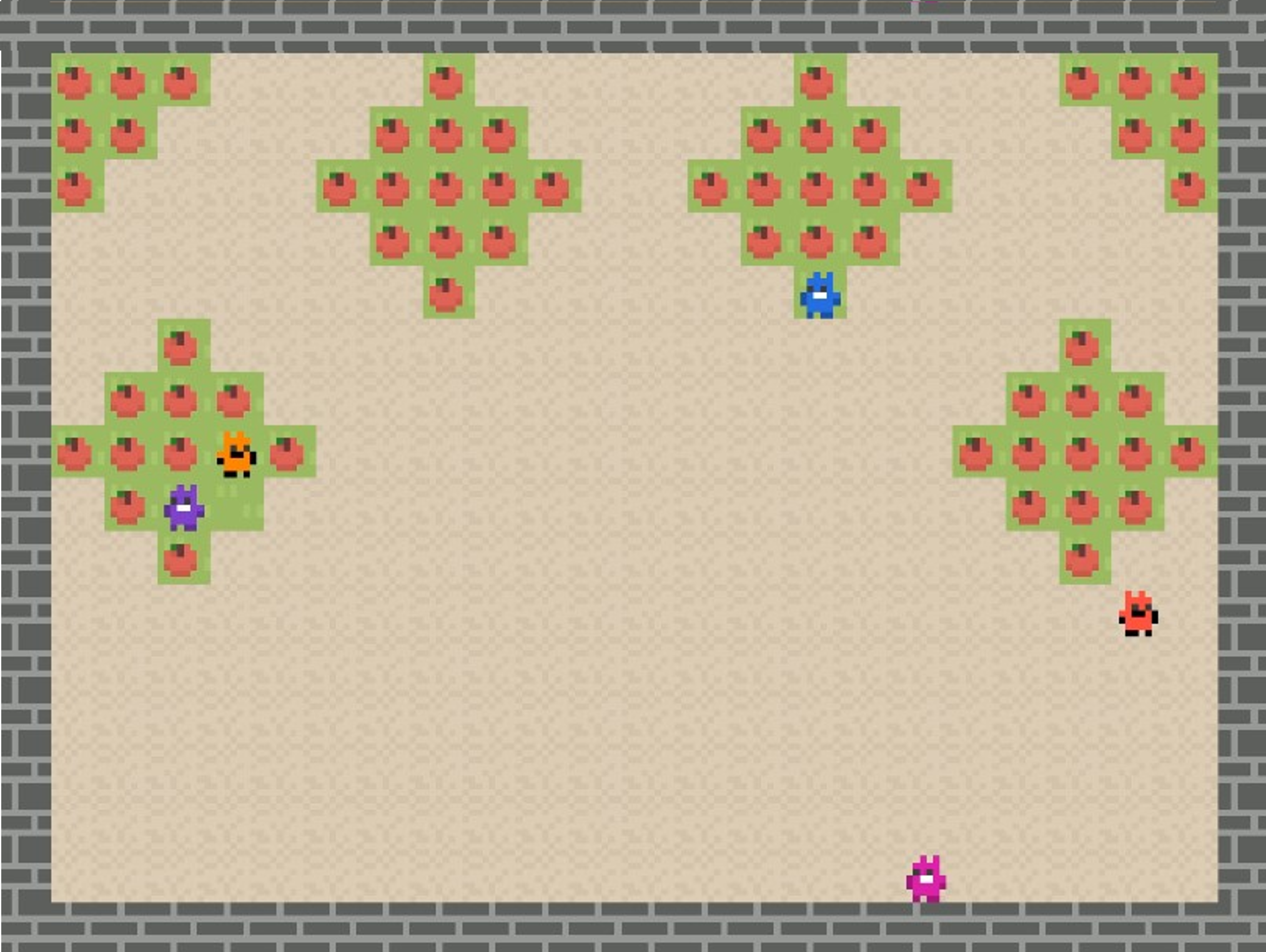}
\caption{This is a screen capture of a running simulation of the Commons Harvest scenario. Bots can be identified by their arms and legs of color black.}\label{environment}
\end{figure}

In the course of our experiments, we selected the ``Commons Harvest'' scenario. In this scenario, agents with unsustainable practices can lead to situations where resources are depleted. This is known as the tragedy of the commons. This scenario is structured around a grid world featuring apples, each conferring a reward of 1 to agents. The regrowth of apples is subject to a per-step probability determined by the apples' distribution in an L2 norm with a radius of 2. Notably, apples may become depleted if there are no other apples in close proximity. Fig. \ref{environment} provides a visual representation of this custom-designed scenario, illustrating the presence of 3 LLM agents and 2 bots.

The LLM agents possess the capacity to execute high-level actions in each round. These actions include: \verb|immobilize player (player_name) at (x, y)|, \verb|go to position (x, y)|, \verb|stay put|, and \verb|explore (x, y)|. They enable the agents to zap other players, navigate to predefined positions on the map, stay in the same position, and explore the world, respectively. On the contrary, bots, characterized as agents trained through reinforcement learning, perform one movement for every two movements made by any of the LLM agents. The policies governing the bots lead them to engage in unsustainable harvesting practices and instigate attacks against other agents in close proximity.

In general, maximizing the welfare of the population for this scenario would require the LLM agents to restrain themselves from eating the last apple on each of the apple trees, and to attack the bots or agents that harvest the apples in an unsustainable way to avoid the depletion of the apples.

\subsubsection{Simulation} In a simulation, each episode of the game involves the participation of a predetermined quantity of LLM agents and bots. The LLM agents take a high-level action on their turn and proceed to execute it until all three LLM agents have completed their respective high-level actions. Meanwhile, the bots are in constant motion, executing a move for every two moves made by any of the agents (note that a high-level action typically comprises more than one movement). The simulation concludes either upon reaching a maximum predetermined number of rounds (typically 100) or prematurely if all the apples in the environment are consumed.

\subsection{Adapting the environment to LLM agents}\label{subsec32}

The Melting Pot scenarios consist of several two-dimensional layers accommodating various objects, each with its own custom logic. While initially, a matrix with distinct symbols seemed the most intuitive way to communicate the game state to the LLMs, it proved challenging for LLMs like GPT-3.5 or GPT-4 to interpret and reason about the spatial information provided by the position of objects in the matrix. To address this issue, we opted to develop an observation generator tailored to this particular environment. In this generator, every relevant object receives a natural language description, supplemented by coordinates expressed as a vector $[x, y]$, denoting row and column respectively. Moreover, some relevant state changes are captured while an agent waits for its turn, and these changes are also captured and communicated to the agents. The complete list of descriptions generated for the objects and events of this environment is shown in Appendix \ref{app1}.

\section{LLM agent architecture}\label{sec4}

The design of the LLM agents predominantly drew upon the Generative Agents architecture \citep{generativeagents}. This choice was motivated by its comprehensive nature, positioning it as one of the most versatile architectures for agents that could be readily tailored to various tasks. While the Voyager architecture \citep{voyager} also presented a viable option, its efficacy was somewhat limited due to its inherent inflexibility. Voyager constructs agent actions dynamically during gameplay, involving the generation and validation of code to execute actions in the environment. In the context of our specific case, it was deemed preferable to externalize actions from the architecture to enhance simplicity.

\begin{figure*}[ht!]
\makebox[\textwidth][c]{%
    \includegraphics[scale=0.18]{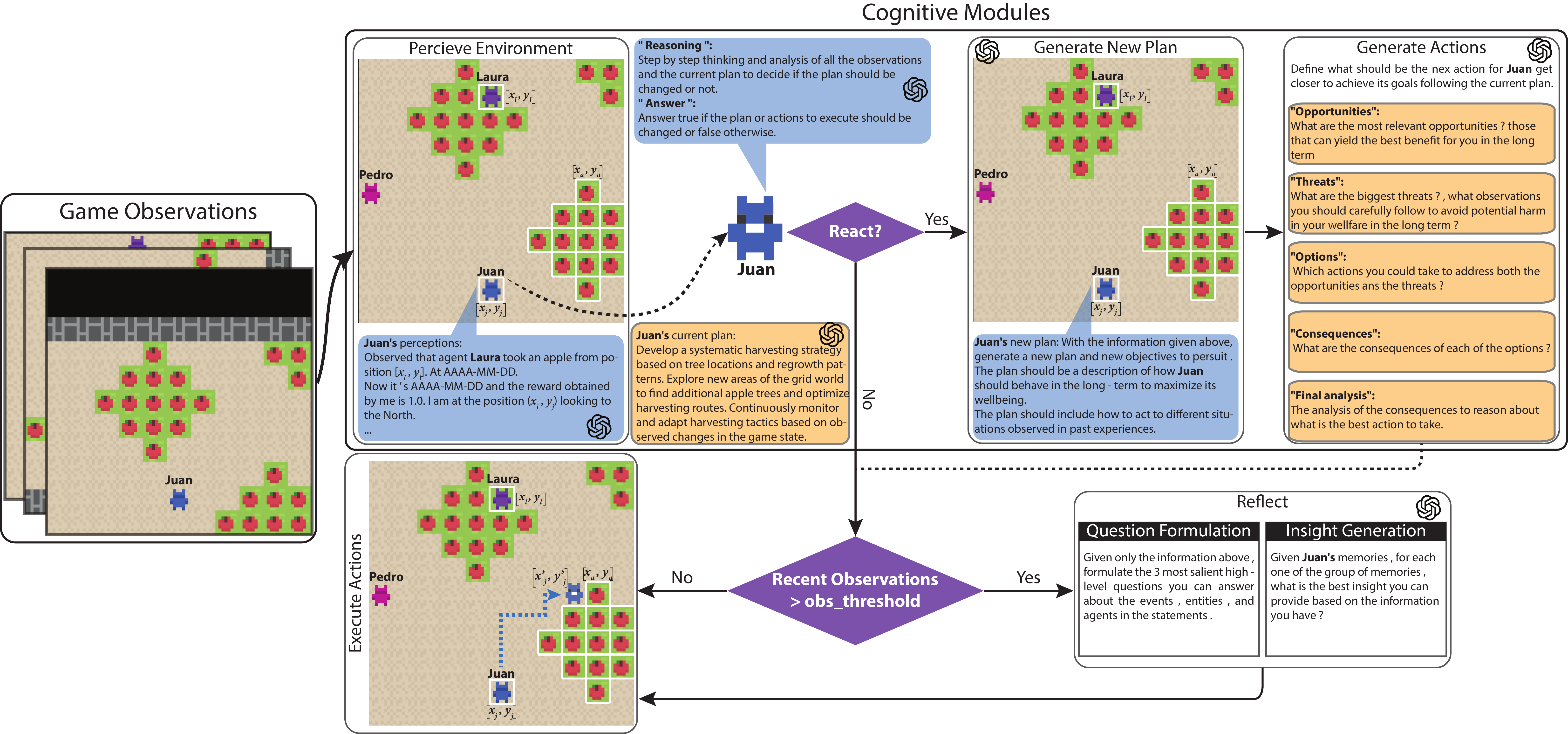}
}
\caption{The flow diagram for an action taken by an LLM agent.}\label{decision-flow_diagram}
\end{figure*}

Fig. \ref{decision-flow_diagram} illustrates the flow diagram outlining the process through which an agent initiates an action. Each action undertaken by LLM agents entails a comprehensive cognitive sequence designed to enhance the agent's reasoning capabilities. This sequence involves the assimilation of feedback from past experiences and the translation of its objectives into a viable plan, enabling the execution of actions within the environment. This architectural framework is in a perpetual state of environmental sensing, generating observations that empower the agent to respond effectively to changes in the world.

\subsection{Memory structures}\label{subsec41}

This agent architecture employs three distinct memory structures designed for specific functions:

\subsubsection{Long-Term Memory} 
This repository stores observations of the environment and various thoughts generated by the agent in its cognitive modules. Leveraging the ChromaDB vector database, memories are stored and the Ada OpenAI model generates contextual embeddings, enabling the agent to retrieve memories relevant to a given query.

\subsubsection{Short-Term Memory} 
To facilitate rapid retrieval of specific memories or information, a Python dictionary is utilized. This dictionary stores information that must always be readily available to the agent, such as its name, as well as data that undergoes constant updates, such as current observations of the world.

\subsubsection{Spatial Memory} 
Given the agent's navigation requirements in a grid world environment, spatial information becomes pivotal. This includes the agent's position and orientation. To support effective navigation from one point to another, utility functions are implemented to aid the agent in spatial awareness and movement.

\subsection{Cognitive modules}\label{subsec42}

\subsubsection{Perception module} 

The initial stage in the cognitive sequence is the Perception Module. This module is tasked with assimilating raw observations from the environment. These observations serve as a comprehensive snapshot of the current state of the world, offering insights into the items within the agent's observable window.

To optimize processing efficiency, the observations undergo an initial sorting based on their proximity to the agent. Subsequently, only the closest observations are channeled to the succeeding cognitive modules. The parameter governing the number of observations passed is denoted as \verb|attention_bandwidth|, initially configured at a value of 10.

Following this, the module undertakes the responsibility of constructing a memory, destined for long-term storage. An illustrative memory example is outlined below:

\begin{lstlisting}[
    language=TeX,
    basicstyle=\footnotesize\ttfamily,
    frame=lines,
    numbers=left,
    numberstyle=\tiny\color{gray},
    linewidth=\textwidth,
    breaklines=true,
    caption=Prompt of the Perceive Module,
    label=act_prompt,
    xleftmargin=2em,
    framexleftmargin=1.5em,
    framexrightmargin=1.5em,
    backgroundcolor=\color{yellow!10}
]
I took the action "grab apple (9, 20)" in my last turn. 
Since then, the following changes in the environment have been observed: 
Observed that agent bot_1 took an apple from position [8, 20]. At 2023-11-19 04:00:00
Observed that agent bot_1 took an apple from position [8, 21]. At 2023-11-19 06:00:00
Observed that an apple grew at position [9, 20]. At 2023-11-19 06:00:00
Observed that agent Laura took an apple from position [2, 15]. At 2023-11-19 07:00:00
Now it's 2023-11-19 09:00:00 and the reward obtained by me is 1.0. I am  at the position (10, 20) looking to the North. 
I can currently observe the following:
Observed an apple at position [9, 20]. This apple belongs to tree 6.
Observed grass to grow apples at position [8, 20]. This grass belongs to tree 6.
\end{lstlisting}

Ultimately, the Perceive module determines whether an agent should initiate a response based on the current observations. During this stage, the agent assesses its existing plan and queued actions to ascertain their suitability. It evaluates whether it is appropriate to proceed with the current course of action or if the observed conditions warrant the development of a new plan and the generation of corresponding actions for execution. The complete prompt is shown in \ref{secA_react_prompt}.

\subsubsection{Planning module} 
This module comes into play once observations have been sorted and filtered. The Planning module leverages the amalgamation of current observations, the existing plan, the contextual understanding of the world, reflections from the past, and rationale to meticulously craft a newly devised plan. This plan intricately outlines the high-level behavior expected from the agent and delineates the goals the agent will diligently pursue. For the complete prompt, refer to \ref{secA_plan_prompt}.

\subsubsection{Reflection module} 
The Reflection module is designed to facilitate profound contemplation on observations and thoughts from fellow agents at a higher cognitive level. Activation of this module is contingent upon reaching a predetermined threshold of accumulated observations. In our experimental setup, reflections were initiated after every 30 perceived observations, roughly translating to three rounds in the game. The Reflection module comprises two key stages:

\begin{enumerate}
\item \textit{Question Formulation:} In the first stage, the module utilizes the 30 retained observations to formulate the three most salient questions regarding these observations.

\item \textit{Insight Generation:} The second stage involves using these questions to retrieve pertinent memories from long-term memory. Subsequently, the questions and retrieved memories are employed to generate three insights, which are then stored as reflections in the long-term memory.
\end{enumerate}

The retrieval of relevant memories employs a weighted average encompassing cosine similarity, recency score, and poignancy scores. The recency score is computed as $e^h$, where $h$ denotes the number of hours since the last memory was recorded. Meanwhile, the poignancy score reflects the intensity assigned to the memory at its point of creation. Throughout the experiments, a uniform poignancy score of 10 was assigned to all memory types. For the complete prompts and more details on question formulation and insight generation processes of this module, refer to Appendix \ref{secA_reflect_prompt}.

\subsubsection{Action Module} This module plays the role of generating an action for the agent to undertake. As detailed in Appendix \ref{secA_act_prompt}, the selection of the action is determined by the Language Model (LLM), which considers the agent's comprehension of the world, its current goals and plans, reflections, ongoing observations, and the available valid actions within the environment. The creation of new action sequences occurs under two conditions: when the current sequence is empty or when the agent is responding to observations. For this prompt, we manually crafted a reasoning structure, similar to those described in Self-Discover \citep{selfdiscover}, to help the LLM consider different alternatives and evaluate them before making the final decision.

\section{Evaluation scenarios}\label{sec5}

To assess the outcomes, we utilized the per capita average reward of the focal population as our primary metric. The focal population comprises LLM agents, and the chosen metric aligns with the Melting Pot framework's approach \citep{meltingpot}, which evaluates population welfare. We compare this metric across two sets of scenarios.

The first set of scenarios is intended to measure how the personality given to the agents affects their welfare. For this purpose, we prepared five scenarios: \begin{inparaenum}[(1)]
    \item as a baseline we do not give the agents any personality specifications (Without personality),
    \item agents are instructed to be cooperative (All coop.),
    \item agents are instructed to be cooperative and provide a short description of how to be cooperative in the chosen scenario (All coop. with def.),
    \item agents are instructed to be selfish (All selfish),
    \item agents are instructed to be selfish and provide a definition with the expected behavior of someone selfish for the given scenario (All selfish with def.).
\end{inparaenum}

The second set of scenarios is more challenging as competition increases by reducing the number of trees and apples, modifying the agents' initial understanding of the social environment, or by adding other entities to the environment (bots). These changes demand a deeper understanding from the agents and swift reactions to master the scenarios. More concretely, the \begin{inparaenum}[(1)]
     first three scenarios consist of an environment where there are three agents and only one apple tree. Each scenario differs in the personality given to the agents: \item all cooperative, \item all selfish, and \item without personality.

     The last scenario of the second set \item has the same base configuration, but with two agents and two bots, where the bots are reinforcement learning agents trained to harvest unsustainably and attack other agents. These bots are part of scenario 0 of the commons harvest open scenario described in Meltingpot 2.0 \citep{meltingpot}.
\end{inparaenum}

We also add a scenario aimed at demonstrating how the information an agent has about the rest of the agents can influence their behavior. In this scenario, the environment starts with the same number of trees; however, from the beginning of the simulation, each agent is informed that among them, one is acting entirely selfishly, representing a risk due to their unsustainable consumption.

For all the experiments, the agents receive information about the environmental rules. They are aware that the per-step growth probability of apples is influenced by nearby apples and that green patches can be depleted if all apples within them are consumed. However, the agents lack information about what is the optimal policy for each scenario, and are unfamiliar with bots and other situations in the game. The complete world context that is given to the agents is shown in Appendix \ref{secA_world_context}.

Ten simulations for each scenario were conducted where the LLM agents were powered by the GPT-3.5 from the OpenAI API for the majority of modules, and GPT-4 powered the action module. On the other hand, the Ada model was used to create contextual embeddings of the memories. Details of the simulation costs are available in Appendix \ref{secA1}.

\section{Results}\label{sec6}
\subsection{Impact of personality in population welfare}\label{sec61}

\begin{figure}[h]%
\centering
\includegraphics[width=0.8\textwidth]{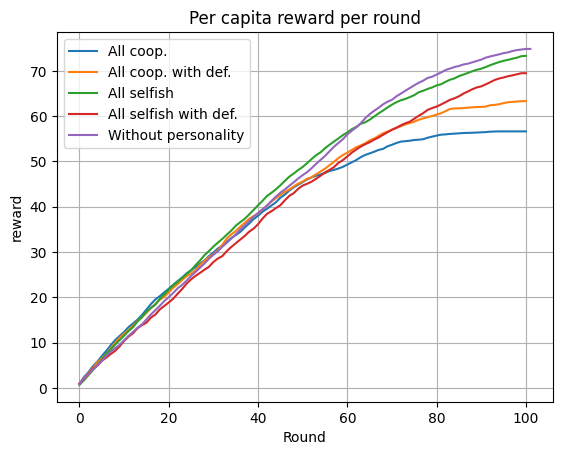}
\caption{The per capita average reward of the agents by scenario. Ten simulations were performed per scenario to assess how the agents' assigned personalities could affect population welfare. The scenario with no particular personality assigned exhibited the best per capita reward, followed by scenarios where agents were instructed to be selfish, and lastly, the worst performance was observed in scenarios where agents were instructed to be cooperative.}\label{set1_avg_rwd}
\end{figure}

\begin{figure}[h]%
\centering
\includegraphics[width=0.85\textwidth]{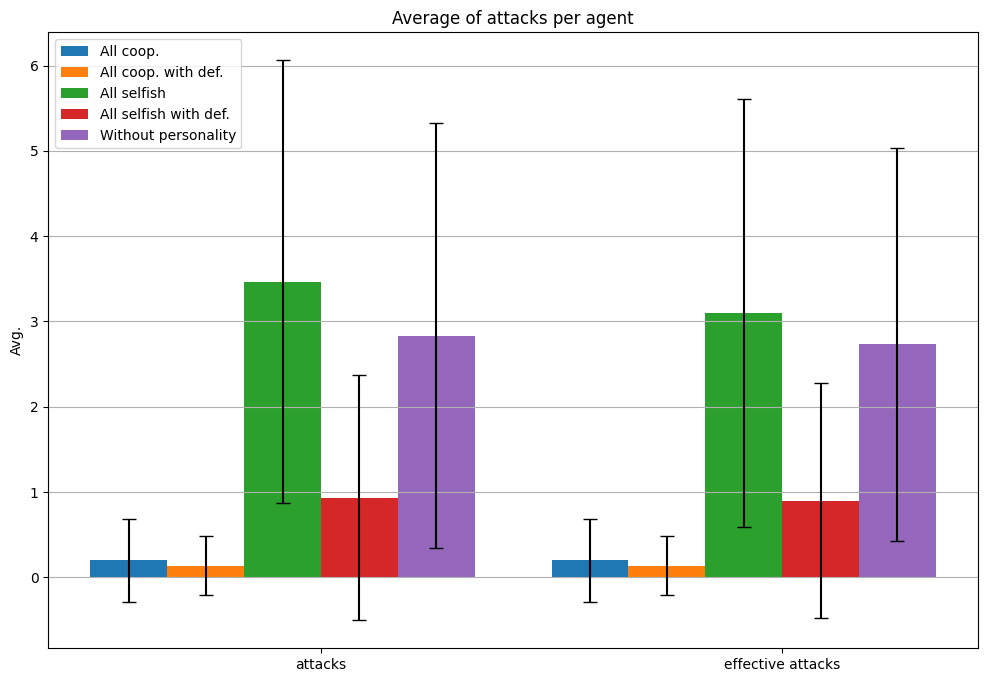}
\caption{The number of times the agents decided to attack and the number of times the attacks were effective, i.e., the number of times the attack hit the other agent, thus removing the agent from the game for the next five moves. The scenarios \textit{All selfish} and \textit{Without personality} registered a higher number of attacks, while the scenarios \textit{All coop.} and \textit{All coop. with def.} showed the least number of attacks.}\label{set1_attacks}
\end{figure}

The average per capita reward obtained for the first set of scenarios is shown in Fig. \ref{set1_avg_rwd}. The best-performing simulations were those where no particular personality description was given to the agents, followed by the scenarios where the agents were instructed to be selfish. Surprisingly, the scenarios where the agents were told to be cooperative had the worst performance. Further analysis revealed that these results are primarily explained mainly by the number of times the agents decided to attack other agents (see Fig. \ref{set1_attacks}).

To gain a better understanding of the agents' behavior, we recorded the number of times the agents decided to attack other agents, and the number of times these attacks were effective. These actions are crucial in the game as they are the only mechanism provided for direct interaction with other agents. They help agents counteract behaviors such as indiscriminate apple picking by other agents, which threatens the depletion of apple trees, or decreasing competition when too many agents are near the same tree. More concretely, when an agent attacks and the ray beam hits its target (another agent), the agent that was hit is taken out of the game for the next five steps and then revived in a random position of the spawning area of the map.

Fig. \ref{set1_attacks} shows the results of these attack indicators for the first set of experiments. The results depict some important differences across the scenarios, mainly reflecting the reluctance of the cooperative agents to attack, and an unexpected difference between the number of attacks of the selfish agents instructed with definition and the selfish agents without definition.

LLMs appear to equate cooperation with refraining from attacking, even when attacking may be the only viable strategy to address uncooperative agents. This behavior was the main cause for cooperative instructed agents to achieve the worst average per capita reward. On the other hand, the selfishly instructed agents behave similarly to the agents lacking assigned personalities, suggesting that LLMs partially disregard the personality given and tend to cooperate by harvesting apples sustainably. The notable disparity in attack frequencies between selfish agents with and without definition is intriguing because agents with the selfish definition decided to explore more frequently rather than attack, the reason for that remains a mistery.

Another important behavior to track is the decisions the agents made when they were near the last apple of a tree. Whether they choose to take it or ignore it is a crucial event and highly impactful on the final per capita reward, as there are only six apple trees in the game, and taking the last apple from a tree means that the tree would be depleted and would not produce more apples. For this reason, we created an indicator that counts how many times the agents closed the distance between themselves and the last apple of a tree, divided by how many times the nearest apple to the agent was the last apple of a tree. However, this indicator does not account for situations where the last apple, despite being the closest to the agent, is not visible to the agent because it is outside the observation window of the agent. This limitation could have impacted the observed results.

\begin{figure}[h]%
\centering
\includegraphics[width=0.85\textwidth]{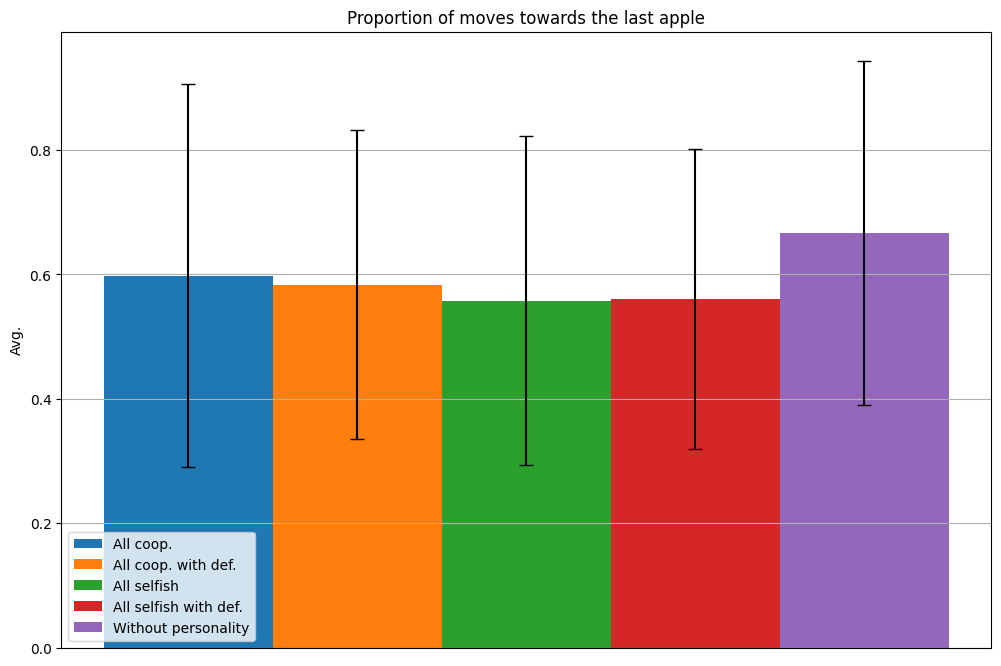}
\caption{Indicator of the number of times the agent closed the distance towards the last apple of a tree divided by the times the last apple of a tree was the nearest to the agent. The results show that there are no important differences between the first set of scenarios.}\label{set1_moves_t_last_apple}
\end{figure}

In Fig. \ref{set1_moves_t_last_apple}, we can see that the proportion of times the agents moved towards the last apple is pretty similar across all the scenarios, indicating that the personality descriptions did not cause a major effect on the awareness of the agents regarding the welfare detriment caused by the depletion of apple trees. These results highlight a limited understanding among the agents regarding the consequences of their actions.

\subsection{Performance of the agents in more challenging scenarios}

The second set of experiments consists of scenarios where the competition increases or the resources become scarcer. The purpose of these scenarios is to measure how the agents respond to the new game conditions.

\subsubsection{One single tree scenarios} The first three scenarios in this set represent an environment involving a more intensive competition for resources. The three agents, who usually have a limited field of vision, are in constant observation of a single tree in the environment, which is situated in a confined space. The difference between each scenario lies in the type of personality assigned to each agent, with the personalities in this case being \textit{All cooperative}, \textit{All selfish}, and \textit{Without Personality}. For practical purposes, no specific definition was given to any personality. The purpose of the scenario is to demonstrate the collective sustainability capacity that different types of agents can have where resources are highly limited.

\begin{figure}[h]%
\centering
\includegraphics[width=1\textwidth]{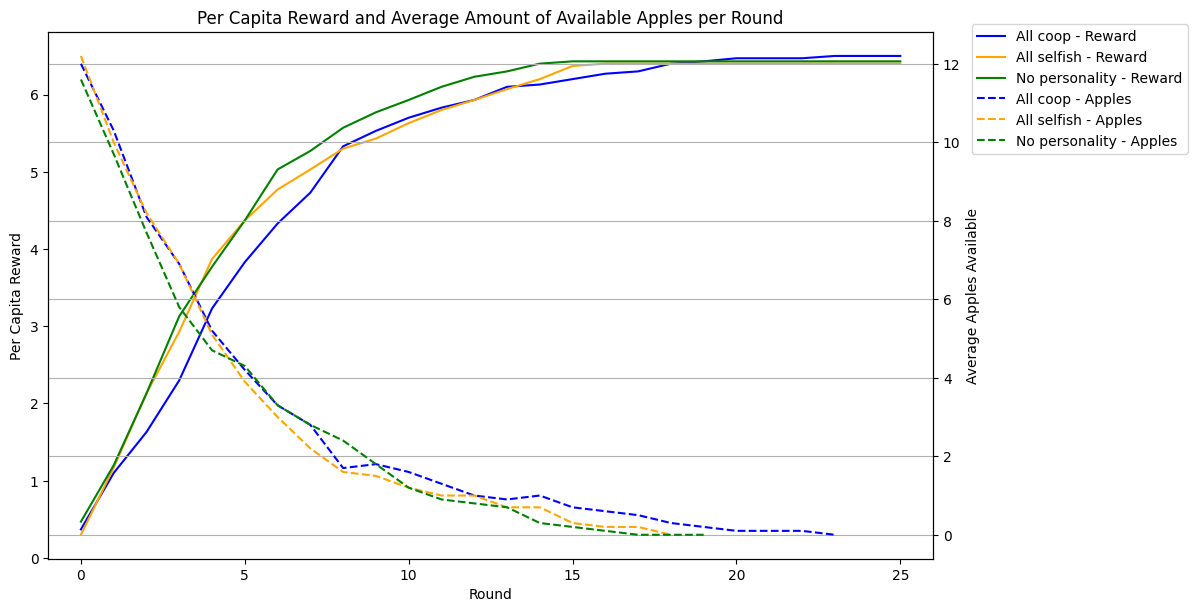}
\caption{Average reward per capita versus average apple availability across personality scenarios when there is only a single tree: The results show a slight superiority in terms of sustainability by cooperative agents, the number of rounds they managed to keep the tree alive was slightly higher than that of the rest of the agents. However, this behavior was not significant enough to obtain a better reward per capita than other agents.}\label{set2_combined_one_tree}
\end{figure}

In Fig. \ref{set2_combined_one_tree}, the results for the ``Per capita reward" are contrasted with the ``Average amount of available apples" for the described group of scenarios. Upon close examination, it is noted that the slope of the reward curve for cooperative agents is less than that for \textit{Selfish} and \textit{Without personality} agents. This behavior contributes to this set of agents having resource availability for a slightly longer period, as shown in the figure. However, given the dynamics of the probability of apple reappearance, this behavior was not significant enough to allow \textit{cooperative} agents to have a considerably superior reward per capita. Therefore, it is concluded that no set of agents was able to demonstrate sufficiently good sustainable behavior due to their lack of understanding of the world and their lack of communication and coordination capabilities with other agents.

\subsubsection{Agents versus Bots}
The fifth scenario of the second set of experiments exposes two agents to the presence of two reinforcement learning bots. The policy of the bots makes them take the apples without regard for the replenishment rate or the risk of depleting the trees; they focus solely on maximizing their rewards by taking the apples, but they also attack other agents, especially where there are no other apples in proximity.

\begin{figure}[h]%
\centering
\includegraphics[width=0.7\textwidth]{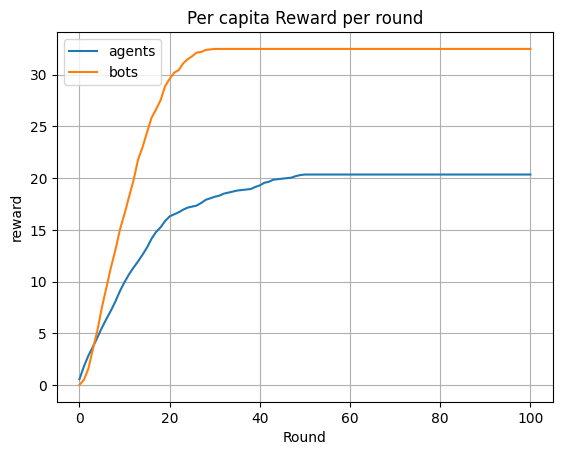}
\caption{Average reward per capita by sub-population (agents and bots). In the results, there is a clear gap between the agents and the bots, where the bots can take advantage of the agents by solely focusing on taking apples without worrying about depleting the trees.}\label{set2_rwd_agents_vs_bots}
\end{figure}

In Fig. \ref{set2_rwd_agents_vs_bots} we see the results of the average reward per capita for the agents versus bots scenario. The initial notable observation is that the bots consistently achieve higher rewards than the agents. This phenomenon is mainly explained by the policy of the bots, which prioritizes taking all the visible apples over other actions, while the agents explore the map or move to other positions on the map with higher frequency than the bots. However, it is important to note how the per capita reward for the bots stops increasing earlier than that for the agents, indicating greater difficulty for the bots to increase their rewards when trees are scarce, compared to the agents. Moreover, we found that in half of the simulations, at least one of the agents achieved a better reward than that of a bot, leading us to conclude that sometimes the agents are capable of outperforming the greedy policy of the bots. Upon closer examination, we observed that in those simulations, the agents were able to find apple trees more easily than the bots and that they also tended to attack when another agent or bot was taking apples from the same tree as them.

\begin{figure}[h]%
\centering
\includegraphics[width=0.65\textwidth]{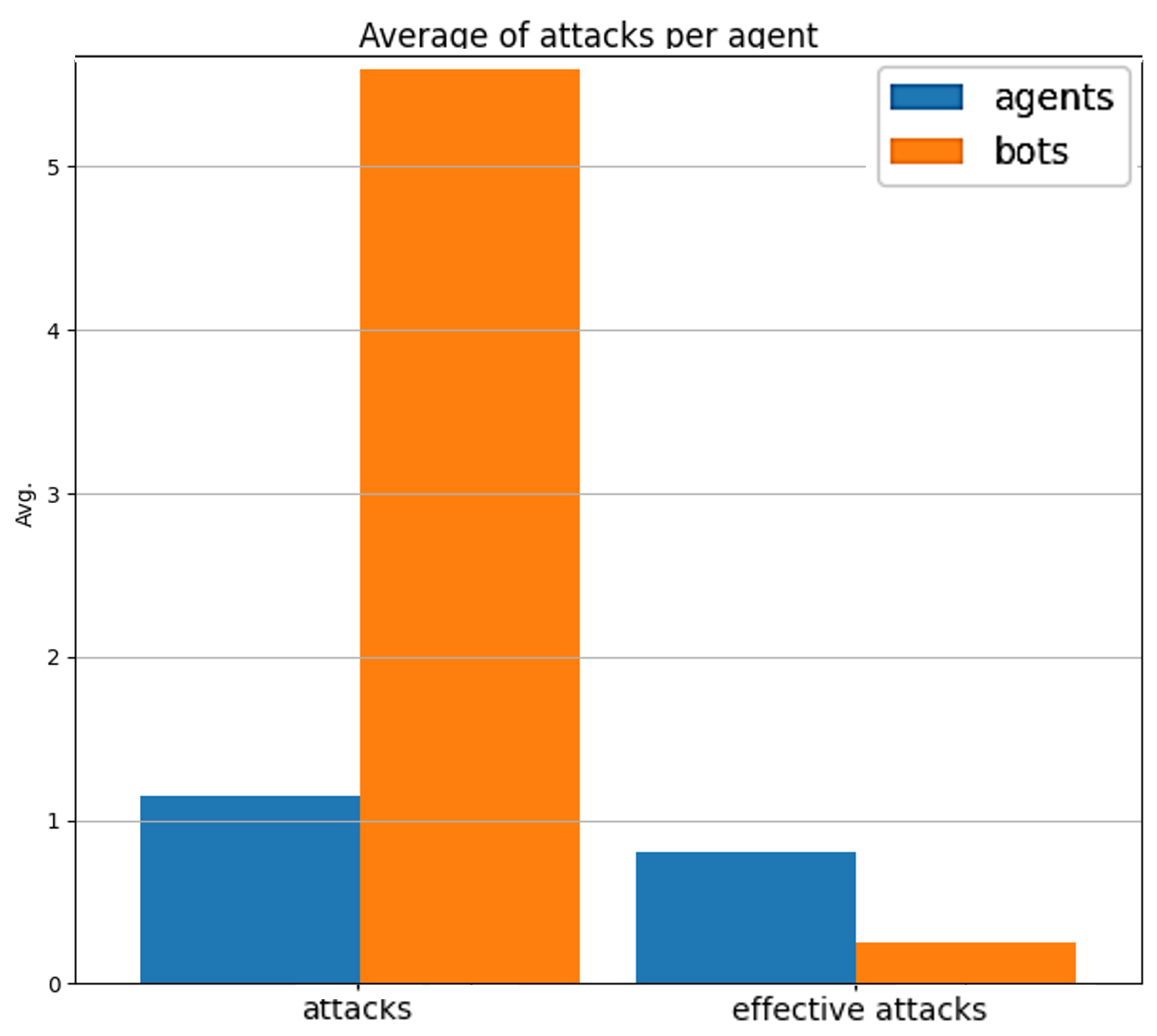}
\caption{The number of times the agents decided to attack and the number of times the attacks were effective. Bots attacked almost five times as frequently as agents. However, the agents' effectiveness was more than double that of the bots.}\label{set2_attacks_agents_vs_bots}
\end{figure}

In Fig. \ref{set2_attacks_agents_vs_bots}, a significant disparity between the number of attacks perpetrated by bots and agents is observed. Despite bots' attacks occurring almost five times as frequently as those executed by agents, the latter proved to be twice as effective in their attacks. Upon manual review of the simulations, we identified that bots increased their frequency of attacks when they were unable to perceive apples within their observation window, even when the attacks were not directed towards any specific target. This finding led us to appreciate how the actions taken by the agents are comparatively more coherent than those of the bots. Furthermore, the behavior of the agents exhibited closer resemblance to human behavior, not only in terms of attacks but also in their movement patterns, in contrast to the seemingly random and redundant actions of the bots.

\begin{figure}[h]%
\centering
\includegraphics[width=0.7\textwidth]{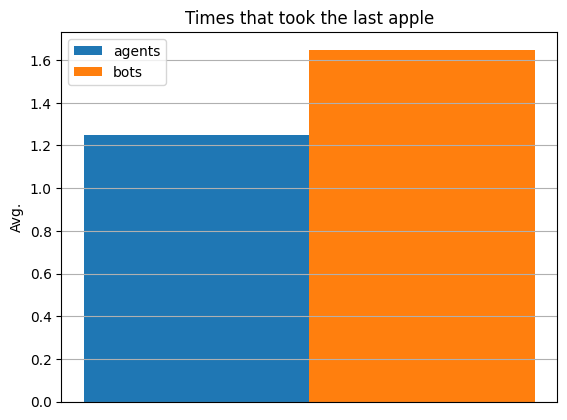}
\caption{Average number of times the agents and bots took the last apple of a tree by sub-population (agents and bots). In the results, we observed that the agents depleted trees less frequently than the bots did, showcasing that the bots were more responsible for the depletion of resources and had a higher negative impact in the population welfare.}\label{set2_took_last_apple_agents_vs_bots}
\end{figure}

Moreover, Fig. \ref{set2_took_last_apple_agents_vs_bots} shows that the agents depleted trees with higher frequency than the bots. Thus, the agents demonstrated the capacity to sometimes restrain themselves from just taking apples by trying to maximize their long-term rewards, whereas bots always prioritized their short-term rewards.

\subsection{Impact of knowledge of other agent's behavior}\par
This experiment considers the hypothetical scenario in which all agents are previously informed that one specific agent is entirely selfish and the implications that its uncooperative behavior can have. Likewise, this agent is informed to act selfishly, providing the previously described definition of selfishness. The objective of this scenario is to highlight the behavior that agents can exhibit when possessing valuable information about their social environment.

\begin{figure}[h]%
\centering
\includegraphics[width=0.65\textwidth]{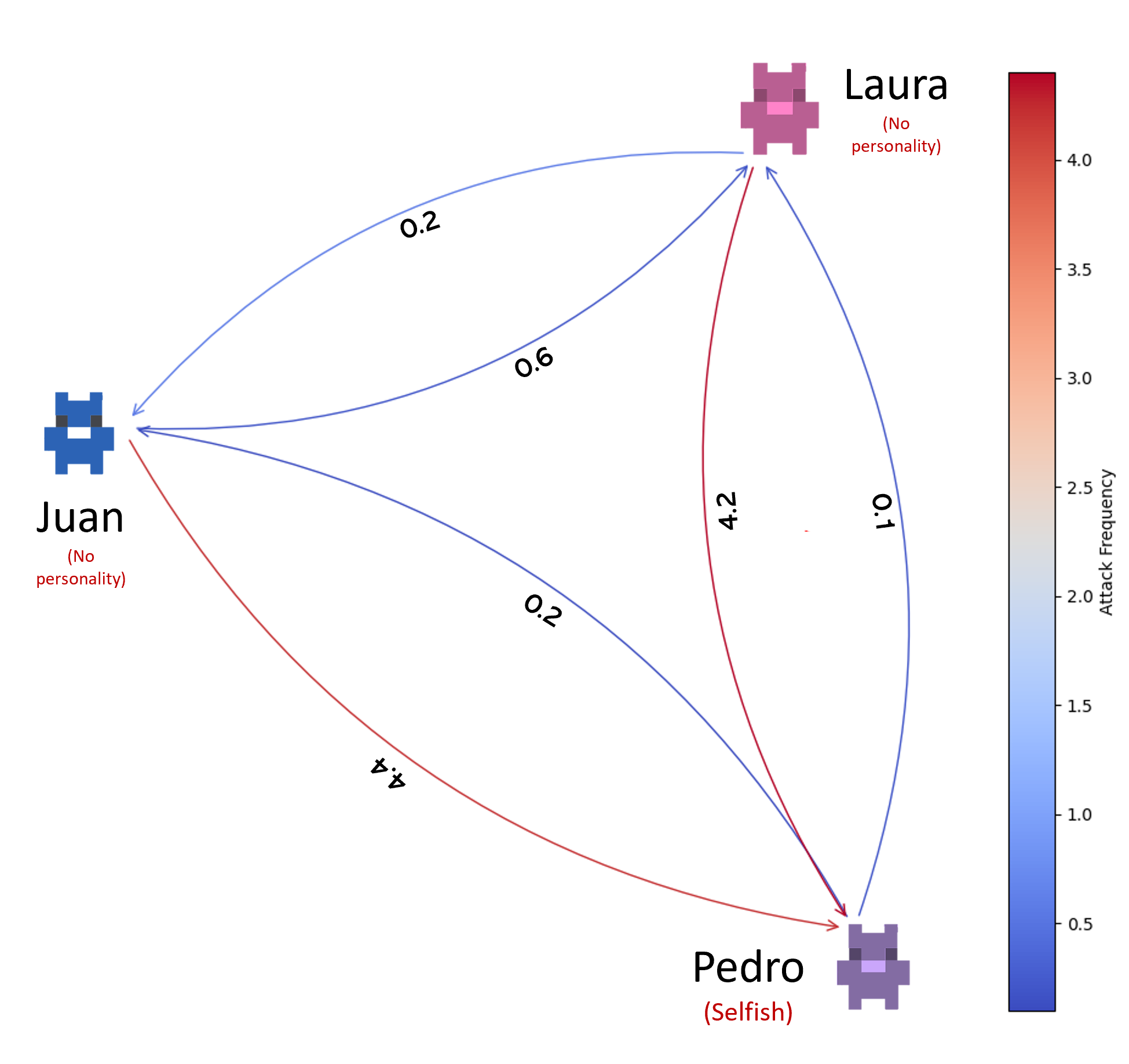}
\caption{Graph depicting the average number of times an agent effectively attacked another agent in the scenario where all agents are informed that ``Pedro" is a ``Selfish agent." At first glance, the results clearly show how the other two agents without personality choose to immobilize Pedro repeatedly throughout the simulations, directing more than 80\% of their attacks exclusively at ``Pedro".}\label{set2_attacks_graph___pedro_is_selfish}
\end{figure}

Fig. \ref{set2_attacks_graph___pedro_is_selfish} shows that, on average, agents without personality targeted Pedro, the selfish agent, exclusively in 86\% of the attacks. This illustrates how the two agents without a defined personality utilized the information forcibly implanted in them to benefit the overall sustainability of the environment, as they repeatedly immobilized the agent who posed a risk due to his excessive consumption and selfish actions. This demonstrates the necessity for agents to acquire this type of information, whether independently through their observations, reflections, and understanding of the world, or through communication with another agent who has previously synthesized this information from their experiences.

\section{Discussion}\label{sec7}

\subsection{Importance of Cooperative Capabilities}\label{subsec71}

In the presented scenarios, experiments detailed in Section \ref{sec3} revealed that the used agent architecture yielded suboptimal results when confronted with unfamiliar situations or when the LLM knowledge couldn't decisively guide optimal decision-making. Furthermore, while agents demonstrated a willingness to cooperate, their actions did not reflect a clear understanding of how to effectively collaborate within the given environment.

To address the proposed scenarios in a better way, agents needed to recognize certain principles. For instance, they should refrain from harvesting the last apple in a green patch to prevent depletion and should engage in cooperation with other agents while avoiding collaboration with the bots or uncooperative agents. Observing that the bots consistently harvested apples unsustainably, agents should have deduced that attacking the bots was necessary to protect the green patches from depletion. This ability to prioritize long-term and collective welfare over short-term rewards, as well as recognizing the divergent behavior and preferences of other entities (bots), aligns with what \citet{openproblems} refer to as cooperative capabilities.

This prompts a consideration of whether current agent architectures genuinely enable cooperative behavior, and if the absence of such capabilities hinders their ability to navigate more intricate tasks and environments. \citet{openproblems} succinctly categorize cooperative capabilities into four essential components:

\begin{enumerate}
\item \textit{Understanding:} Agents must comprehend the world, anticipate the consequences of their actions, and demonstrate an understanding of the beliefs and preferences of others.

\item \textit{Communication:} Vital for achieving understanding and coordination, communication should be intentional, serving as a tool to gather information and coordinate efforts. Agents should be equipped to assess the intentions of others and establish their own criteria for discerning relevant information. Moreover, agents do not always have common interests, the other agent could be trying to deceive or convince in its self-interest.

\item \textit{Commitment:} Cooperation is often hindered by commitment problems arising from an inability to make credible promises or threats. Agent architectures should address these issues by providing mechanisms for agents to enforce or establish credibility in their promises and threats.

\item \textit{Institutions:} Social structures, such as institutions, play a crucial role in simplifying interactions between agents. These structures define the rules of the game for all entities, potentially extending to the allocation of roles, power, and resources.
\end{enumerate}

In essence, cultivating collaborative capabilities within agent architectures is crucial for tackling the complexities inherent in diverse tasks and environments. Historically, agent architectures have inadequately endowed agents with such capabilities. Instances such as Generative Agents \citep{generativeagents} and the Improving Factuality and Reasoning in Language Models through Multiagent Debate \citep{MultiAgentDebate} enable agents to engage in conversations or observe the perspectives of others. However, these approaches are hampered by the absence of independent evaluation criteria and discernment specific to the current limitations of LLMs.

\subsection{Cooperative Agent Architecture}\label{subsec72}

\begin{figure}[h]%
\centering
\includegraphics[width=\textwidth]{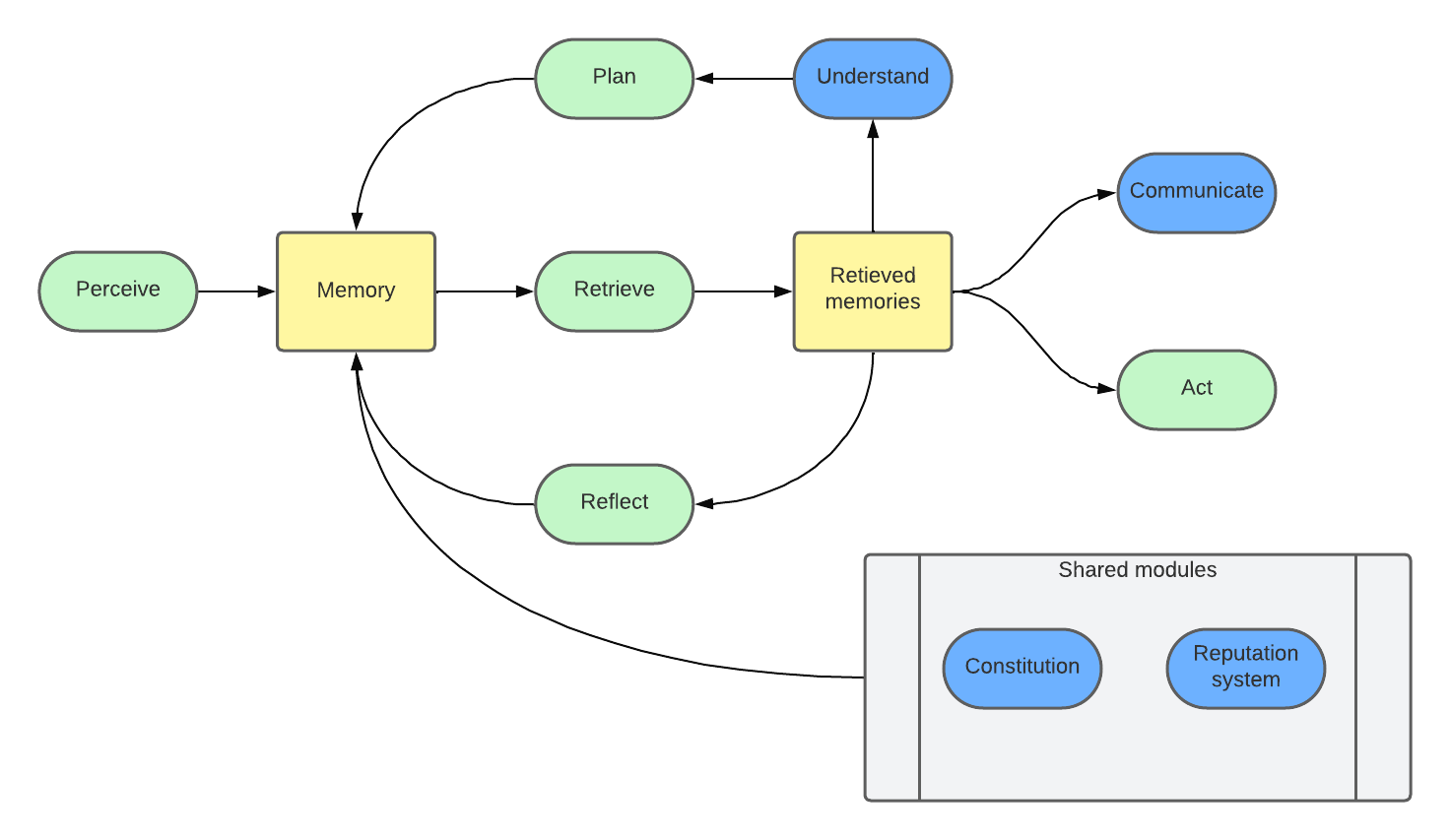}
\caption{Diagram of the proposed cooperative architecture. The modified or new modules are painted in blue.}\label{coop_arch}
\end{figure}

Based on previous findings, we propose an architecture to enhance agents' cooperative capabilities (see Fig. \ref{coop_arch}). In this architecture, several new modules are proposed:

\begin{enumerate}

\item \textit{Understanding module:} This component is tasked with a comprehensive analysis of the agent's memories, fostering a deeper comprehension of the surrounding world. The agent's proficiency extends to predicting the behaviors of fellow agents and discerning environmental changes, enabling it to take actions with a keen awareness of their potential consequences. Notably, the agent must possess the capacity to infer both the governing principles of the world and the underlying motivations guiding others' actions. This inference capability extends to scenarios where these principles may deviate from common knowledge or the pre-training model knowledge. \citet{learnrules} demonstrate that LLMs, like GPT-4, can learn such rules when explicitly prompted to identify them, utilizing question-answer pairs to later apply the learned rules in problem-solving.
The proposed module operates by initially extracting the rules and behavioral patterns of the world and other agents. It achieves this by prompting the LLM with historical world observations and the current state of the world, aiming to identify rules that explain the current state based on the agent's observations. These identified rules are initially stored as world hypotheses. As the agent utilizes these hypotheses to interpret the current state, they are transformed into explicit rules once they surpass a predefined threshold. Additionally, the LLM is prompted to generate predictions about future states of the environment, empowering the agent to make informed decisions guided by anticipated future scenarios.

\item \textit{Communication module:} The primary objective of this module is to equip the agent with the ability to engage in intentional communication with other agents. Two key objectives have been identified to enhance cooperative capabilities: (1) The agent is encouraged to seek new information from other agents. It must decide whether there are pertinent questions that can be posed to fellow agents, aiding in a better understanding of the world or gaining insights into the preferences of others. This information is pivotal for augmenting the agent's overall comprehension. (2) Agents are provided with the opportunity to negotiate and establish agreements deemed mutually beneficial. These agreements are stored in memory in a specialized manner to hold agents accountable for their commitments. The goal is to foster improved coordination among agents, thereby enhancing collaborative efforts.

\item \textit{Constitution Module:} This module plays a crucial role in establishing a shared foundation for all agents. Its primary function is to define a set of common rules, providing agents with an initial framework to comprehend the world and formulate assumptions about the behavior of other agents. The constitution also delineates the consequences, whether penalties or rewards, that agents may face for specific behaviors or interactions. This not only lends credibility to agreements among agents but also discourages undesirable behaviors, streamlining interactions and cultivating a cooperative environment.

\item \textit{Reputation System:} This system is designed to hold agents accountable for their actions. It evaluates each agent based on their adherence to agreements made with other agents. Periodically, the system prompts a language model with the existing agreements and corresponding actions, requesting a reputation score. This score is then accessible to all agents, influencing communication dynamics and aiding in understanding the behavior of others. Additionally, it facilitates making predictions about future states.

\end{enumerate}

\section{Conclusion and Future Work}\label{sec8}

Cooperative capabilities have been somewhat overlooked in LLMs' agent architectures, yet they may represent the crucial element enabling agents to accomplish pioneering tasks and thrive in intricate environments. As large language models (LLMs) advance, agent architectures stand to gain significantly by attaining enhanced responses from LLMs, particularly in tasks demanding substantial reasoning or when confronted with copious information in the prompt.

In this paper, our objective is to ascertain whether LLMs-enhanced autonomous agents can operate cooperatively. To this end, we adapt the Melting Pot scenarios to textual representations that can be easily operationalized by LLMs, and implement a reusable architecture for the development of LAAs employing the modules proposed in Generative Agents \citep{generativeagents}. This architecture includes short and long-term memories, and cognitive modules of perception, planning, reflection, and action. The ``Commons Harvest'' game was used to test the resulting system, and the results were evaluated from the viewpoint of cooperative metrics in different proposed scenarios. 

The results indicate a gap in the current agents' cooperative capabilities vis-à-vis unfamiliar situations. Agents showed a cooperative tendency but lacked an adequate understanding of how to collaborate effectively in an unknown environment. The agents needed to understand complex factors like the need to conserve resources, identify non-cooperative agents, and prioritize collective welfare over short-term gains. The research thereby draws attention to the need for a more inclusive architecture fostering cooperation and enhancing agent capabilities, including superior understanding, effective communication, credible commitment, and well-defined social structures or institutions.

Responding to the findings, we also proposed to improve the architecture with several modules to enhance the cooperative capabilities of the agents. These include an understanding module responsible for a comprehensive analysis of the agent's memory and surroundings, a communication module to enable intentional information exchange, a constitution module that lays out common rules of engagement, and a reputation system that holds agents accountable for making decisions for the collective good. Our future efforts will be focused on building and evaluating this cooperative architecture.

\section*{Data Availability}

All the data generated for each simulation and the summary files for each experiment are available at \href{https://zenodo.org/records/11221750}{experiments data}. The code repository will be shared on GitHub upon acceptance of the paper.

\section*{Acknowledgments}

This work is supported by Google through the Google Research Scholar Program.

\newpage

\appendix

\section{Descriptions generated for the objects in the environment}
\label{app1}

In Table \ref{tab_objs_description}, we show all the natural language descriptions generated to represent the relevant objects and events of the Commons Harvest scenario of Melting Pot.

\begin{table}[H]
\caption{Natural language description by object or event}\label{tab_objs_description}
\centering
\begin{tabular*}{\textwidth}{@{\extracolsep{\fill}} p{0.2\textwidth} p{0.8\textwidth}}
\toprule
\textbf{Object/Event} & \textbf{Description} \\
\midrule
Other agent  & Observed agent \verb|<agent_name>| at position \verb|[<x>, <y>]|. \\
Grass  & Observed grass to grow apples at position \verb|[<x>, <y>]|. This grass belongs to tree \verb|<tree_id>|. \\
Apple & Observed an apple at position \verb|[<x>, <y>]|. This apple belongs to tree \verb|<tree_id>|. \\
Tree & Observed tree \verb|<tree_id>| at position \verb|[<x>, <y>]|. This tree has \verb|apples_number| apples remaining and \verb|grass_number| grass for apples growing on the observed map. The tree might have more apples and grass on the global map. \\
Observed someone being attacked & Someone was attacked at position \verb|[<x>, <y>]|. \\
Observed a ray beam & Observed a ray beam from an attack at position \verb|[<x>, <y>]|. \\
Observed an apple was taken & Observed that agent \verb|agent_name| took an apple from position \verb|[<x>, <y>]|. \\
Observed grass disappeared & Observed that the grass at position \verb|[<x>, <y>]| disappeared. \\
Observed grass grew & Observed that grass to grow apples appeared at position \verb|[<x>, <y>]|. \\
Observed apple grew & Observed that an apple grew at position \verb|[<x>, <y>]|. \\
The agent was attacked & There are no observations: You were attacked by agent \verb|agent_name| and currently you're out of the game. \\
The agent is out of the game & There are no observations: you're out of the game. \\
\bottomrule
\end{tabular*}
\end{table}

\section{Knowledge about the world given to agents}\label{secA_world_context}\label{app2}
The Listing \ref{world_context_prompt} shows the raw world description passed to the agents. This is the only information agents have about the environment.

\begin{lstlisting}[
    language=TeX,
    basicstyle=\footnotesize\ttfamily,
    frame=lines,
    numbers=left,
    numberstyle=\tiny\color{gray},
    linewidth=\textwidth,
    breaklines=true,
    caption=World Context given to agents,
    label=world_context_prompt,
    xleftmargin=2em,
    framexleftmargin=1.5em,
    framexrightmargin=1.5em,
    backgroundcolor=\color{yellow!10}
]
I am in a misterious grid world. In this world there are the following elements:
Apple: This object can be taken by any agent. The apple is taken when I go to its position. Apples only grow on grass tiles. When an apple is taken it gives the agent who took it a reward of 1.
Grass: Grass tiles are visible when an apple is taken. Apples will regrow only in this type of tile based on a probability that depends on the number of current apples in a L2 norm neighborhood of radius 2. When there are no apples in a radius of 2 from the grass tile, the grass will disappear. On the other hand, if an apple grows at a determined position, all grass tiles that had beeen lost will reappear if they are between a radius of two from the apple.
Tree: A tree is composed from apples or grass tiles, and it is a tree because the patch of these tiles is connected and have a fix location on the map. These trees have an id to indentify them.
Wall: These tiles delimits the grid world at the top, the left, the bottom, and the right of the grid world. 
The grid world is composed of 18 rows and 24 columns. The tiles start from the [0, 0] position located at the top left, and finish on the [17, 23] position located at the bottom right.
I am an agent and I have a limited window of observation of the world.
\end{lstlisting}

\section{React Prompt}\label{secA_react_prompt}

The Listing \ref{react_prompt} shows the entire prompt used in the react module. This prompt enables the agent to decide whether to react to the current observations—where reacting implies altering the plan and generating a new action. The prompt receives inputs in the following order: name, world context, current observations, current plan, actions to take if any, changes observed in the game state, game time, and agent's personality.

\begin{lstlisting}[
    language=TeX,
    basicstyle=\footnotesize\ttfamily,
    frame=lines,
    numbers=left,
    numberstyle=\tiny\color{gray},
    linewidth=\textwidth,
    breaklines=true,
    caption=Prompt of the Perceive Module,
    label=react_prompt,
    xleftmargin=2em,
    framexleftmargin=1.5em,
    framexrightmargin=1.5em,
    backgroundcolor=\color{yellow!10}
]
You have this information about an agent called <input1>:

<input6>

<input1>'s world understanding: <input2>

Current observations at <input7>:
<input3>

<input6>

<input8>

Current plan: <input4>

Actions to execute: <input5>

Review the plan and the actions to execute, and then decide if <input1> should continue with its plan and the actions to execute given the new information that it's seeing in the observations.
Remember that the current observations are ordered by closeness, being the first the closest observation and the last the farthest one.

The output should be a markdown code snippet formatted in the following schema, including the leading and trailing "```json" and "```", answer as if you were <input1>:

```json
{
 "Reasoning": string, \\ Step by step thinking and analysis of all the observations and the current plan to decide if the plan should be changed or not
 "Answer": bool \\ Answer true if the plan or actions to execute should be changed or false otherwise
}
\end{lstlisting}

\section{Plan prompt}\label{secA_plan_prompt}

The Listing \ref{plan_prompt} shows the raw prompt used in the plan module. This prompt helps the agent make a high-level plan and define several goals to guide its actions. The inputs that this prompt receives are the following in order: name, world context, current observations, current plan, reflections, reason to react,  agent's personality, and changes observed in the game state.

\begin{lstlisting}[
    language=TeX,
    basicstyle=\footnotesize\ttfamily,
    frame=lines,
    numbers=left,
    numberstyle=\tiny\color{gray},
    linewidth=\textwidth,
    breaklines=true,
    caption=Prompt of Planning Module,
    label=plan_prompt,
    xleftmargin=2em,
    framexleftmargin=1.5em,
    framexrightmargin=1.5em,
    backgroundcolor=\color{yellow!10}
]
You have this information about an agent called <input1>:

<input7>

<input1>'s world understanding: <input2>

Recent analysis of past observations:
<input5>

Observed changes in the game state:
<input8>

Current observations: 
<input3>

Current plan: <input4>
This is the reason to change the current plan: <input6>

With the information given above, generate a new plan and new objectives to persuit. The plan should be a description of how <input1> should behave in the long-term to maximize its wellbeing.
The plan should include how to act to different situations observed in past experiences.

The output should be a markdown code snippet formatted in the following schema, including the leading and trailing "```json" and "'''", answer as if you were <input1>:

```json
{
 "Reasoning": string, \\ Step by step thinking and analysis of all the observations and the current plan to create the new plan and the new goals.
 "Goals": string, \\ The new goals for <input1>.
 "Plan": string \\ The new plan for <input1>. Do not describe specific actions.
}'''
\end{lstlisting}

\section{Reflection prompts}\label{secA_reflect_prompt}

The Listing \ref{reflect_prompt1} shows the raw prompt used in the first part of the reflections module i.e. question formulation. The inputs for this prompt are the following: name, world context, accumulated observations since the last reflection, and agent's personality.

The prompt used in the insight generation part that takes place in the reflect module is shown in Listing \ref{reflect_prompt2}, its corresponding inputs are the following: name, world context, group of memories retrieved for each generated question in the first part, and agent's personality.

\begin{lstlisting}[
    language=TeX,
    basicstyle=\footnotesize\ttfamily,
    frame=lines,
    numbers=left,
    numberstyle=\tiny\color{gray},
    linewidth=\textwidth,
    breaklines=true,
    caption=Prompt of  Reflect Module for question formulation,
    label=reflect_prompt1,
    xleftmargin=2em,
    framexleftmargin=1.5em,
    framexrightmargin=1.5em,
    backgroundcolor=\color{yellow!10}
]
You have this information about an agent called <input1>:

<input4>

<input1>'s world understanding: <input2>

Here you have a list of statements:
<input3>

Given only the information above, formulate the 3 most salient high-level questions 
you can answer about the events, entities, and agents in the statements. 


The output should be a markdown code snippet formatted in the following schema, 
including the leading and trailing "```json" and "'''", answer as if you were <input1>:

```json
{
    "Question_1": {
        "Reasoning": string \\ Reasoning for the question 
        "Question": string \\  The question itself
    },
    "Question_2": {
        "Reasoning": string \\ Reasoning for the question 
        "Question": string \\ The question itself
    },
    "Question_3": {
        "Reasoning": string \\ Reasoning for the question
        "Question": string \\ The question itself
    }
}'''
\end{lstlisting}

\begin{lstlisting}[
    language=TeX,
    basicstyle=\footnotesize\ttfamily,
    frame=lines,
    numbers=left,
    numberstyle=\tiny\color{gray},
    linewidth=\textwidth,
    breaklines=true,
    caption=Prompt of Reflect Module for insights generation,
    label=reflect_prompt2,
    xleftmargin=2em,
    framexleftmargin=1.5em,
    framexrightmargin=1.5em,
    backgroundcolor=\color{yellow!10}
]
You have this information about an agent called <input1>:

<input4>

<input1>'s world understanding: <input2>

Here you have a list of memory statements separated in groups of memories:
<input3>

Given <input1>'s memories, for each one of the group of memories, what is the best insight you can provide based on the information you have? 
Express your answer in the JSON format provided, and remember to explain the reasoning behind each insight.

The output should be a markdown code snippet formatted in the following schema, 
including the leading and trailing "```json" and "'''", answer as if you were <input1>:

```json
{
    "Insight_1": {
        "Reasoning": string \\ Reasoning behind the insight of the group of memories 1
        "Insight": string \\ The insight itself
    },
    "Insight_2": {
        "Reasoning": string \\ Reasoning behind the insight of the group of memories 2
        "Insight": string \\ The insight itself
    },
    "Insight_n": {
        "Reasoning": string \\ Reasoning behind the insight of the group of memories n
        "Insight": string \\ The insight itself
    }
}'''
\end{lstlisting}

\section{Act prompt}\label{secA_act_prompt}

The Listing \ref{act_prompt} shows the raw prompt used in the act module. This prompt is in charge of deciding which action to take. The inputs that this prompt receives are the following: name, world context, current plan, the most recent ten reflections, current observations, number of actions to generate, set of valid actions, current goals, agent's personality, position of the known trees, portion of the map explored, previous actions, and changes observed in the game state.

\begin{lstlisting}[
    language=TeX,
    basicstyle=\scriptsize\ttfamily,
    frame=lines,
    numbers=left,
    numberstyle=\tiny\color{gray},
    linewidth=\textwidth,
    breaklines=true,
    caption=Prompt of Action Module,
    label=act_prompt,
    xleftmargin=2em,
    framexleftmargin=1.5em,
    framexrightmargin=1.5em,
    backgroundcolor=\color{yellow!10}
]
You have this information about an agent called <input1>:

<input10>

<input1>'s world understanding: <input2>

<input1>'s goals: <input9>

Current plan: <input3>

Analysis of past experiences: 
<input4> 

<input11>

Portion of the map explored by <input1>: <input12>

Observed changes in the game state:
<input14>

You are currently viewing a portion of the map, and from your position at <input6> you observe the following: 
<input5>

Define what should be the nex action for Laura get closer to achieve its goals following the current plan.
Remember that the current observations are ordered by closeness, being the first the closest observation and the last the farest one.
Each action you determinate can only be one of the following, make sure you assign a valid position from the current observations and a valid name for each action:

Valid actions: 
<input8>

Remember that going to positions near the edge of the portion of the map you are seeing will allow you to get new observations.
<input13>

The output should be a markdown code snippet formatted in the following schema, including the leading and trailing "```json" and "'''", answer as if you were Laura:
```json
{
    "Opportunities": string \\ What are the most relevant opportunities? those that can yield the best benefit for you in the long term
    "Threats": string \\ What are the biggest threats?, what observations you should carefully follow to avoid potential harm in your wellfare in the long term?
    "Options: string \\ Which actions you could take to address both the opportunities ans the threats?
    "Consequences": string \\ What are the consequences of each of the options?
    "Final analysis: string \\ The analysis of the consequences to reason about what is the best action to take
    "Answer": string \\ Must be one of the valid actions with the position replaced
}'''
\end{lstlisting}

\section{Simulations Cost}\label{secA1}

\begin{table}[htbp]  
\caption{Costs of simulations}\label{tab1}
\centering
\renewcommand{\arraystretch}{1.5} 
\begin{tabular}{@{} lcc @{}}  
\toprule
 & \textbf{Avg. Simulation} & \textbf{Avg. Execution} \\
\textbf{Experiment} & \textbf{Cost (\$)} & \textbf{Time (minutes)} \\
\midrule
Set 1 - No bio  & $8.57 (0.93)$ & $151.18 (17.04)$ \\
Set 1 - All Coop  & $7.00 (1.58)$ & $119.50 (26.78)$ \\
Set 1 - All Coop with def\footnotemark[1]  & $15.63 (5.69)$ & $212.60 (65.14)$ \\
Set 1 - All Selfish  & $8.60 (1.84)$ & $127.63 (26.57)$ \\
Set 1 - All Selfish with def  & $9.73 (1.59)$ & $217.50 (65.57)$ \\
Set 2 - One tree - no bio  & $0.78 (0.28)$ & $16.41 (6.94)$ \\
Set 2 - One tree - all coop  & $0.78 (0.17)$ & $13.93 (2.99)$ \\
Set 2 - One tree - all selfish  & $0.83 (0.30)$ & $13.72 (4.38)$ \\
Set 2 - Agents vs. Bots  & $1.88 (0.66)$ & $27.86 (11.83)$ \\
Set 3 - All aware one selfish  & $10.05 (0.88)$ & $151.93 (10.56)$ \\
\bottomrule
\end{tabular}
\footnotetext[1]{For this experiment, the models of OpenAir were used through Classic.}
\end{table}

\newpage

\bibliographystyle{elsarticle-harv} 
\bibliography{bibliography}



\end{document}